# Accurate and Efficient Multi-Channel Time Series Forecasting via Sparse Attention Mechanism

1st Lei Gao
*Department of Machine Learning*
*SF Express*
Shanghai, China
gaolei1@sf-express.com

2nd Hengda Bao
*Department of Machine Learning*
*SF Express*
Shanghai, China
baohengda@sf-express.com

3rd Jingfei Fang
*Department of Machine Learning*
*SF Express*
Shanghai, China
fangjingfei@sf-express.com

4th Guangzheng Wu
*Department of Management*
*Zhejiang University of Technology*
Hangzhou, China
ouroboros2035@163.com

5th Weihua Zhou
*Department of Management*
*Zhejiang University*
Hangzhou, China
larryzhou@zju.edu.cn

6th Yun Zhou
*Department of Management*
*Zhejiang University of Technology*
Hangzhou, China
zhouyun@zjut.edu.cn

***Abstract*—The task of multi-channel time series forecasting is ubiquitous in numerous fields such as finance, supply chain management, and energy planning. It is critical to effectively capture complex dynamic dependencies within and between channels for accurate predictions. However, traditional method paid few attentions on learning the interaction among channels. This paper proposes Linear-Network (Li-Net), a novel architecture designed for multi-channel time series forecasting that captures the linear and non-linear dependencies among channels. Li-Net dynamically compresses representations across sequence and channel dimensions, processes the information through a configurable non-linear module and subsequently reconstructs the forecasts. Moreover, Li-Net integrates a sparse Top-K Softmax attention mechanism within a multi-scale projection framework to address these challenges. A core innovation is its ability to seamlessly incorporate and fuse multi-modal embeddings, guiding the sparse attention process to focus on the most informative time steps and feature channels. Through the experiment results on multiple real-world benchmark datasets demonstrate that Li-Net achieves competitive performance compared to state-of-the-art baseline methods. Furthermore, Li-Net provides a superior balance between prediction accuracy and computational burden, exhibiting significantly lower memory usage and faster inference times. Detailed ablation studies and parameter sensitivity analyses validate the effectiveness of each key component in our proposed architecture.**

***Keywords*—Multivariate Time Series Forecasting, Sparse Attention Mechanism, Multimodal Information Fusion, Non-linear relationship**

## I. Introduction

Time series forecasting is a central task in many fields including finance [1], supply chain management [2], and energy planning[3]. Although single-channel forecasting has been extensively studied, many real-world scenarios involve multi-channel time series data, where multiple interrelated channels evolve simultaneously. Effectively exploiting the interdependencies between these channels is crucial to achieve accurate and robust prediction[4]. For example, in the financial field [5], price changes of multiple currencies and macroeconomic indicators can be jointly used for portfolio risk prediction; in the traffic field, the flow status of different streets can be used to the predict the traffic volume [6].

Attention mechanism has become a powerful tool for capturing the complex dynamic relationships within and among these time series [12]. Attention-based algorithms allow them to adaptively balance the importance of different temporal historical information, demonstrating superior capability of capturing long-range dependencies and subtle temporal patterns, significantly outperforming traditional recurrent or convolutional architectures [13].

However, the application of attention mechanisms may incur a huge computational burden [12]. The computational complexity of self-attention operations increases quadratically with the length of the sequence $O(L^2)$, which is computationally expensive for processing long sequences common in practical applications. This limitation inspires the development of efficient sparse attention strategies [14]. By selectively focusing on the most relevant subset of input elements, the sparse attention mechanism aims to approximate the effect of full attention at an affordable computational cost, making it key for scalable time series models [15].

In addition to the temporal dynamics within the channel, another dimension of complexity stems from the rich auxiliary information that is often available in real-world systems. Integrating such multimodal information - such as calendar features (e.g., holidays, weekends), semantic embeddings of product categories, and real-time operational status - can provide critical situational signals and greatly improve forecasting accuracy [16]. However, how to effectively integrate these heterogeneous data modalities with the original time series data is still a core challenge and opportunity to improve the performance of multi-channel prediction models.

In this paper, we introduce Li-Net, an framework specifically designed for accurate and efficient multi-channel time series prediction. Li-Net addresses the above challenges by integrating the sparse top-K softmax attention mechanism into a multi-scale projection framework. Our framework dynamically condense the sequence and channel dimensions, processes the information through a nonlinear module (MLP or Transformer), and subsequently reconstructs the prediction. The core innovation is the ability to seamlessly incorporate and fuse multimodal embeddings to guide the sparse attention process to focus on the most informative time steps and feature channels. Through extensive experiments on real-world datasets, we demonstrate that Li-Net achieves

Lei Gao, Hengda Bao and Jingfei Fang contributed equally to this work.

competitive performance compared to state-of-the-art baseline algorithms from Tsinghua University Time Series Library [17], providing an excellent balance between accuracy and computational efficiency.

The specific contributions of this paper can be summarized as follows:

- We propose a multi-scale projection mechanism based on Top-K sparse Attention: To overcome the quadratic computational complexity problem faced by traditional attention mechanisms on long sequences, we design a dynamic and learnable sparse attention mechanism. Through Top-K approximation, the proposed mechanism adaptively achieves feature compression and focusing in two dimensions of time and channel, which significantly reduces the computational and memory of the algorithm while maintaining the ability to capture long-range dependencies.
- A unified multi-modal information fusion framework is designed: the framework effectively integrates heterogeneous information from multiple sources, including raw time series data, date embeddings, concept embeddings, store embeddings, and real-time sales status. These multi-modal features are jointly input into the gating mechanism to guide the sparse attention process to more accurately identify the most critical information for prediction, so as to comprehensively improve the representation ability and context understanding level of the framework.
- We build a flexible and efficient general framework for multi-channel time series forecasting: The overall design of Li-Net employs an Encoder-Decoder structure and integrates a configurable feature variation module. This module supports replacement with MLP or Transformer encoding layers, which enables the framework to balance expressive power and computational efficiency according to the requirements and complexity of different tasks. This design enhances the generality and practicability of the framework.

Full experimental validation on real datasets: We comprehensively evaluate Li-Net on several publicly available time series benchmark datasets. Experimental results show that compared with a series of state-of-the-art baseline algorithms in TSLib, Li-Net achieves competitive performance in prediction accuracy, while showing significant advantages in computational efficiency. Detailed ablation experiments further verify the effectiveness of each key component in the Li-Net framework.

The remainder of this paper is structured as follows: Chapter 2 reviews related work; Chapter 3 formalizes the problem definition. Chapter 4 covers the design of Li-Net in detail. Chapter 5 shows the experimental results and analysis. Finally, Chapter 6 concludes the paper.

## II. Related Work

### A. Multi-channel time series prediction

The core challenge of multi-channel time series prediction is how to effectively model the dynamic dependencies between variables. Traditional methods such as LSTM often regard each channel as an independent sequence, ignoring the potential association between them [18]. In recent years, researchers have proposed new solutions mainly from two aspects: architecture design innovation and modeling perspective transformation.

The Transformer proposed by Google Research introduces attention mechanism for the first time [11]to adaptively learn the dependence between each channel, which lays a foundation for subsequent research. CrossFormer makes an important innovation at the architectural level, proposing Dimension-Segment-Wise (DSW) Embedding[19], which decouples time and channel axis: Local patterns are captured by piecewise embedding on the timeline, learnable router vectors to efficiently aggregate and distribute global dependencies across dimensions, The two-stage attention mechanism is used to realize cross-dimensional modeling with controllable complexity. Similarly, iTransformer inverts the structure of traditional transformers, converts single-channel sequences into tokens, and uses cross-channel attention instead of temporal attention to further optimize the efficiency of multivariate joint modeling [20].

From the modeling perspective, SOFTS proposes the Centralized Stochastic Pooling Method to compress multi-channel features into a global core representation, which is embedded and spliced with each channel and fused by MLP to achieve efficient feature interaction with linear complexity [21]. TSMixer directly adopts the pure MLP architecture and learns the importance weight of each channel through the cross-channel coordination head, which proves that the channel association can be effectively modeled without attention mechanism[22]. In addition, Connecting the Dots is pioneering in viewing multivariate time series as graph structures, adaptively learning sparse adjacency matrices, and alternately performing graph convolution (spatial modeling) and temporal convolution (temporal modeling), which is a pioneer in perspective transformation [23]. FourierGNN further abandons the traditional time series module and only processes time series information in the graph domain through Fourier Transform, realizing the end-to-end prediction completely based on graph structure [24]. Lstm-fcns proposed MLSTM-FCN and MALSTM-FCN, which introduced squeee-and-excitation module to enhance multivariate interaction modeling, and combined LSTM and fully convolutional network to achieve efficient time series classification [25].

### B. Sparse attention mechanism

The Transformer algorithm has an $O(n^2)$ complexity bottleneck, which has triggered a large amount of research on the sparse attention mechanism. Sparse Transformers propose Strided Attention and Fixed Attention, reducing the complexity to $O(n\sqrt{n})$ [26]. BigBird integrates Global Attention, Sliding Window Attention and Random Attention, reducing the complexity to O(n) while maintaining the expressiveness of the model, and has become a benchmark for long sequence modeling [27]. Reformer introduces LSH Attention and, in combination with reversible residual layers, supports 64K ultra-long sequence training for the first time [15]. Sparsemax outputs sparse distributions through Euclidean projection [28]. SparseGen proposed a λ/q parameterization scheme to control sparsity [29]. MultiMax innovatively uses the piecewise modulation scheme to solve the sparsity contradiction of SoftMax in multimodal tasks[30]. Top-KAST achieves dynamic sparsity during gradient update through the dual-mask mechanism for forward and backward passes [31]. Differentiable Top-K operators provide strict mathematical guarantees for sparse training [32].

### *C. Multi-modal information fusion*

Multimodal fusion is gradually replacing single-modal modeling and becoming a key direction for improving the accuracy and interpretability of time series prediction [33]. By introducing auxiliary information such as text and images, especially by leveraging the powerful prior knowledge of multimodal large models, time series prediction models can achieve higher-order reasoning that is closer to human cognition [34]. This technological evolution has mainly gone through three stages: from feature concatenation, deep interaction to a unified prediction framework.

In the early exploration, the research mostly adopted simple feature concatenation methods. CNN-BiLSTM directly concatenates the temporal features extracted by BiLSTM with the text features processed by CNN [35]. Probabilistic PV Forecasting parses the weather text through GPT-Agent, converts it into structured numerical features, and then combines it with sensor data [36].

Subsequently, the focus of the research shifted to the deep interaction between modes. To more fully explore the correlations among modalities, researchers have designed more complex interaction mechanisms. The Modality-aware Transformer specifically learns the deep interaction between text and timing signals by introducing the modality-aware encoder and Inter-modal MHA[37]. MoAT, on the other hand, utilizes attention weights to dynamically perform weighted fusion on the prediction results from different modalities[38].

Nowadays, cutting-edge research is beginning to construct a new paradigm of joint prediction and unified modeling. The latest work is no longer content with the utilization of auxiliary information but attempts to build a more integrated framework. Multi-Modal Forecaster proposes a hybrid framework that can simultaneously predict text events and numerical sequences [39]. The CITAB model extracts temporal and text features respectively through CNN-InceptionV2 and TextCNN, and then concatenates them into a multimodal feature set after optimization by the attention mechanism. Finally, carbon emission prediction is achieved by BiLSTM, with significant improvements in both multimodal fusion and prediction accuracy [40].

## III. Problem Statement

In this section, we introduce the problem to be solved, while defining the necessary related notations.

**Definition 1 (Time Series)** A complete set of observation time series for N channels at T steps:

$$\boldsymbol{Y} = \{\boldsymbol{Y_1}, \boldsymbol{Y_2}, \dots, \boldsymbol{Y_N}\} \in \mathbb{R}^{N\times T}$$

Where $\boldsymbol{Y_i} = \{Y_{i,1}, Y_{i,2}, \dots, Y_{i,T}\}$ is the observation time series of the ith channel. When N=1, the time series is a single-channel time series. When N > 1, this time series will become a multi-channel time series, and there is a potential dependence between the channels.

**Definition 2 (Multimodal Time Series)** Time series with multimodal information that can be used for predicting observed values:

$$\boldsymbol{X} = \{\boldsymbol{X_1}, \boldsymbol{X_2}, \dots, \boldsymbol{X_N}\} \in \mathbb{R}^{N\times T\times M}$$

**Algorithm 1 Embedding Module**

**Input:** Historical date $\mathbf{e_{HD}}$, Future date $\mathbf{e_{FD}}$, Store ID $\mathbf{e_S}$, Item ID $\mathbf{e_I}$

**Output:** Historical date embedding vector $\mathbf{V^{HD}}$, Future date embedding vector $\mathbf{V^{FD}}$,Store embedding vector $\mathbf{V^{S}}$,Item embedding vector $\mathbf{V^{I}}$

1:**for** each $\mathbf{e_i}$ in **E do**

2: $\mathbf{H_k} \leftarrow \mathbf{BERT}(\mathbf{Tokenize}(\mathbf{e_i}))$ k represents the number of tokens that $\mathbf{e_i}$ has been split into.

3: $\mathbf{V^i}$←AVGPOOLING(H)

4:**end for**

5:**for** every two $\mathbf{v^i}$, $\mathbf{v^j}$ in **V do**

6: Similarity matrix←compute $\mathbf{cos}(\mathbf{v^i}, \mathbf{v^j})$

7: compute CosSENT Loss function

8: adjust Similarity matrix

9:**end for**

10:**return** $\mathbf{V^{HD}}, \mathbf{V^{FD}}, \mathbf{V^{S}}, \mathbf{V^{I}}$

Here, $\boldsymbol{X_i} = \{\boldsymbol{X_{i,1}}, \boldsymbol{X_{i,2}}, \dots, \boldsymbol{X_{i,M}}\} \in \mathbb{R}^{N\times T}$ denotes the multimodal information matrix of the i^th channel, which incorporates temporal, operational and channel-specific information, etc., with a multimodal dimensionality of M.

**Definition 3 (Multi-Channel Time Series Forecasting)** The task of multichannel time series forecasting is to train a model using a subset $\boldsymbol{D^1} = \boldsymbol{X^1} \cup \boldsymbol{Y^1}$ and then perform forecasting on another subset $\boldsymbol{D^2} = \boldsymbol{X^2} \cup \boldsymbol{Y^2}$:

$$\boldsymbol{X^1} = \{\boldsymbol{X_1^1}, \boldsymbol{X_2^1}, \dots, \boldsymbol{X_N^1}\} \in \mathbb{R}^{N\times t_1 \times M}, t_1 < T$$

$$\boldsymbol{Y^1} = \{\boldsymbol{Y_1^1}, \boldsymbol{Y_2^1}, \dots, \boldsymbol{Y_N^1}\} \in \mathbb{R}^{N\times t_1}, t_1 < T$$

$$\boldsymbol{X^2} = \{\boldsymbol{X_1^2}, \boldsymbol{X_2^2}, \dots, \boldsymbol{X_N^2}\} \in \mathbb{R}^{N\times t_2 \times M}, t_2 < T$$

$$\boldsymbol{Y^2} = \{\boldsymbol{Y_1^2}, \boldsymbol{Y_2^2}, \dots, \boldsymbol{Y_N^2}\} \in \mathbb{R}^{N\times t_2}, t_2 < T$$

Here, $\boldsymbol{Y_i^1}$ represents the historical observed time series of the $i^{th}$ channel, while $\boldsymbol{Y_i^2}$ denotes its "future" observed time series, with their lengths satisfying $t_1 + t_2 = T$. Furthermore, to perform forecasting, we need to learn a model $\mathcal{M}$ that outputs the predictions $\hat{Y}$. This process can be defined as:

$$\boldsymbol{\hat{Y}^2} = \mathcal{M}(\boldsymbol{X^2}) \in \mathbb{R}^{N\times t_2}$$

## IV. Design of Li-Net

The core goal of the Li-Net neural network framework is to deal with sequence data with spatio-temporal characteristics or high-dimensional feature dependencies. Through the cognitive process of "encoding-processing-decoding", the original high-dimensional input sequence is compressed in the time and feature dimensions, and the core calculation and pattern learning are carried out in an abstract low-dimensional space. Finally, the results are reconstructed and mapped to the final prediction target. The whole process deeply integrates multi-modal information, and guides the flow and aggregation of information through dynamically generated attention weights.

As shown in Fig 1, Li-Net can be divided into four modules. We complete the vectorization of text information and the stitching of multi-modal vectors in the multi-modal embedding module. Subsequently, the feature encoding work is completed in the encoding module in a "time-channel" order.

In the channel encoding process, the model is allowed to perceive part of the information of the future target at the encoding stage, so as to generate a more guided and predictive internal representation. The highly abstract and compressed internal representation will be input into the nonlinear transformation module to complete feature conversion, and finally the processed features will be mapped back to the decoding module to complete the output of the predicted value. In the encoding module and the decoding module, for the screening of features, we choose the mechanism of Top-K Softmax dynamic generation of attention to adaptively determine the information that needs to be focused on in the time and feature dimensions. This design can clearly see which key time features and commodity features the model focuses on in the encoding and decoding process. As a result, the model can be more interpretable and intelligently capture key patterns and dependencies in the sequence.

*A. Multimodal Embedding module*

In this paper, multi-modal information[1] including structured data and text data is introduced, so multi-modal information embedding is needed to realize the fusion of various information. The core of the module is to apply text2vec model to convert the original text sequence $W = [w_1, w_2, \ldots, w_L]$ into structured vector data $v = \text{text2vec}(W) \in \mathbb{R}^d$, so as to realize text embedding. Where d is the vector dimension after embedding.

*1) The backbone network of text2vec*

text2vec is an open source text embedding model specially designed for Chinese text processing. It can convert Chinese sentences or short paragraphs into high-dimensional dense vectors, which can capture the semantic information of the text and be used for various natural language processing tasks.

In the actual training, the acquisition and training of sentence vectors depend on the BERT model. Sentences are first segmented by tokenizer according to character or subword granularity, and each word segmentation is converted into a series of Token vectors by BERT, and then these tokens are averaged into a fixed sentence vector.

In the training phase, BERT is used as the initialization backbone network, only the top layers are fine-tuned, and the CoSENT loss function is used to optimize the ordering of sentence vectors in the cosine space. The core goal of the loss function is to bring semantically similar sentences closer in the "direction" of the vector space, rather than only in numerical value. It learns to make the cosine similarity between semantically similar sentences as large as possible, and the cosine similarity between semantically dissimilar sentences as small as possible.

BERT tokenizes a sentence and converts each tokenization into a series of Token vectors. This process can be formalized as follows.

$$\mathrm{H} = \mathrm{Transformer}(\mathrm{Tokenize}(\mathrm{sent})) \in \mathbb{R}^{\mathrm{L} \times \mathrm{D}}$$

Where sent represents the input sentence, L is the length of the token sequence, D is the hidden layer dimension of the model, and H is the vector embedding matrix of the output.

Then, these tokens are merged into a fixed sentence vector by mean pooling, that is, the average of the token embedding vectors in the sentence is calculated as the sentence vector, and the mathematical expression is as follows.

[1]It is worth mentioning that in other scenarios, any multimodal information can be incorporated.

**Algorithm 2 Feature Encoder**

**Input:** Original time series $\mathbf{T}_{\boldsymbol{O}}$, Historical date embedding tensor $\mathbf{T_{HD}}$, Store embedding tensor $\mathbf{T_s}$, Item embedding tensor $\mathbf{T_I}$, Time feature compression level $\mathbf{L_T}$,Channel feature compression level $\mathbf{L_C}$,
**Output:** Encoded feature tensor $\mathbf{T_e}$ of shape [B, C', T']
1:T' ←Compute(X.C//$\mathbf{L_C}$)
2:C' ←Compute(X.T//$\mathbf{L_T}$)
3: Logits-1$\in \mathbb{R}^{B\times T\times T'}$ ←MLP(Concatenate($\mathbf{T_d}, \mathbf{T_s}$))
4:$\mathbf{T_{te}} \in \mathbb{R}^{B\times T\times T'}$ ←**TopKSoftmax**(Logits-1)
5:$\mathbf{T_t} \in \mathbb{R}^{B\times C\times T'} \leftarrow \mathbf{T}_{\boldsymbol{O}} \cdot \mathbf{T_{te}}$
6:$\mathbf{T_{ce-1}} \in \mathbb{R}^{B\times C\times C'}$ ←MLP(Concatenate($\mathbf{T_t}$))
7:$\mathbf{T_{ce-2}} \in \mathbb{R}^{B\times C\times C'}$ ←MLP($\mathbf{T_I}$))
8:$\mathbf{T}_{\boldsymbol{ce}}$←**TopKSoftmax**($\mathbf{T_{ce-1}} \oplus \mathbf{T_{ce-2}}$)
9: $\mathbf{T_e}$ ←Transposition($\mathbf{T}_{\boldsymbol{ce}}$)· $\mathbf{T_t}$
10:**return** $\mathbf{T_e}$

$$\mathrm{e_{sentence}} = \frac{1}{\mathrm{L}}\sum_{\mathrm{i}=1}^{\mathrm{L}} \mathrm{H_i} \tag{1}$$

Here, $\mathrm{H_i}$ denotes the embedding vector of the ith Token.

*2) Loss function for text2vec*

The CoSENT model is an improved scheme proposed on the basis of Sentence-BERT, and its core goal is to solve the "training-prediction inconsistency" problem in SBERT. From CoSENT, we can see that since we end up using cosine similarity to measure sentence similarity in the prediction stage, we should directly optimize this cosine similarity in the training stage instead of learning it indirectly through an intermediate task.

To this end, CoSENT designs a new loss function, which directly uses cosine similarity as the optimization objective, so that the training and prediction objectives of the model are completely consistent. The CoSENT loss function aims to optimize the relative order between all pairs of positive and negative samples within a batch. The mathematical expression is as follows:

$$\mathcal{L} = \log\left(1 + \sum_{(i,j)\in\Omega_{pos},(k,l)\in\Omega_{neg}} v^{\lambda(\cos(u_k,u_l) - \cos(u_i,u_j))}\right) \tag{2}$$

Where $\Omega_{pos}$ is the set of all pairs of positive samples (labeled as similar) in the current batch, and $\Omega_{neg}$ is the set of all pairs of negative samples (labeled as dissimilar) in the current batch. $u_i, u_j, u_k, u_l$ are the embedding vectors for sentences i,j,k,l, respectively, cos(·,·) denotes the cosine similarity function, and $\lambda$ is a hyperparameter used to scale the input of the exponential function to control the strength of the optimization. And the cosine similarity mathematical expression is as follows:

$$\cos(u, v) = \frac{u \cdot v}{\|u\|\|v\|} \tag{3}$$

In the CoSENT loss function, if the cosine similarity of positive sample pairs cos $(u_i, u_j)$ is greater than the cosine similarity of negative sample pairs $\cos(u_k, u_l)$, $v^{negative}$ will be a small number, and its contribution to $\mathcal{L}$ is small. On the contrary, when cosine similarity cos $(u_i, u_j)$ of positive samples is less than cosine similarity $\cos(u_k, u_l)$ of negative samples, $v^{positive}$ will be a number greater than 1, significantly increasing L, resulting in a large gradient to the model and forcing the model to update parameters to correct the error. The external $\log(1+\sum(\cdot))$ structure aggregates the penalty terms of all sample pairs smoothly. It ensures that the loss function responds even if only one sample pair is incorrectly ranked, and that the loss increases monotonically as the number of incorrectly ranked sample pairs increases.

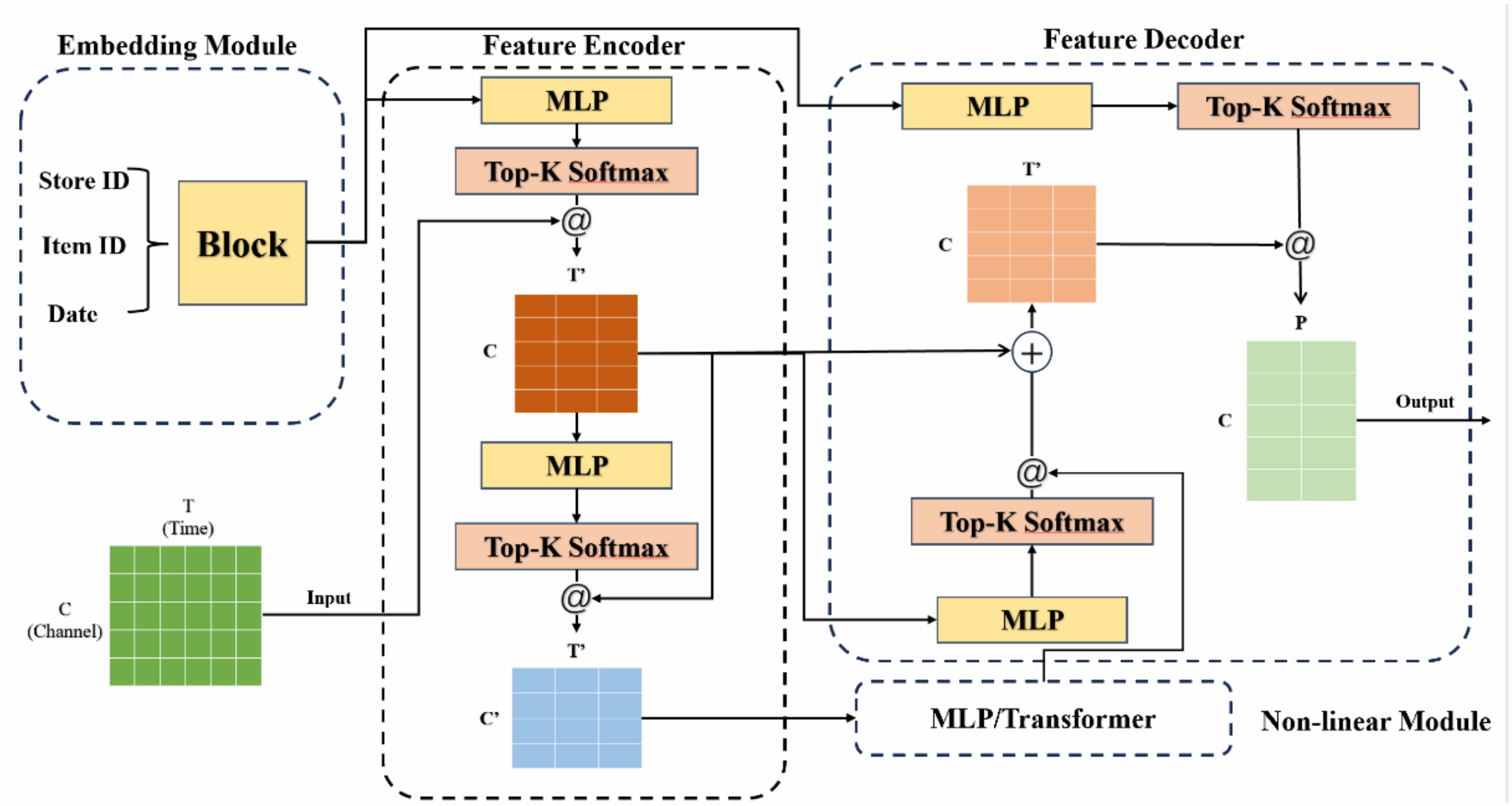

**Fig.1: Overview of Linet**

In the CoSENT loss function, if the cosine similarity of positive sample pairs $\cos(u_i, u_j)$ is greater than the cosine similarity of negative sample pairs $\cos(u_k, u_l)$, $v^{negative}$ will be a small number, and its contribution to $\mathcal{L}$ is small. On the contrary, when cosine similarity $\cos(u_i, u_j)$ of positive samples is less than cosine similarity $\cos(u_k, u_l)$ of negative samples, $v^{positive}$ will be a number greater than 1, significantly increasing L, resulting in a large gradient to the model and forcing the model to update parameters to correct the error. The external $\log(1+\sum(\cdot))$ structure aggregates the penalty terms of all sample pairs smoothly. It ensures that the loss function responds even if only one sample pair is incorrectly ranked, and that the loss increases monotonically as the number of incorrectly ranked sample pairs increases.

In terms of time scales, Li-Net the historical time series date $T_{hd} = [t_1, t_2, \ldots, t_n] \in \mathbb{R}^n$ and the date of the future forecast time step $T_{fd} = [t_{n+1}, t_{n+2}, \ldots, t_{n+p}] \in \mathbb{R}^p$ input text2vec model; In terms of channel scale, Li-Net takes a static variable $S \in \mathbb{R}^{d_{static}}$ as input. Although the input of each scale is different, the resulting embedding vector will be of the same dimension size, which can be formally expressed as $V^{HD}, V^{FD}, V^S, V^I \in \mathbb{R}^{d_{embed}}$.

## B. *Feature encoder*

The feature encoder is the core component of the Li-Net framework. Its goal is to adaptively extract the most significant time and channel features from high-dimensional, multimodal input data and achieve efficient dimensionality reduction. However, introducing too many features makes it difficult to complete the prediction work relatively accurately. On the contrary, it will also introduce noise. Therefore, in order to summarize the patterns of features at the time and channel scales, we use MLP to complete the encoding and compression of features, and through the Top-K Softmax attention mechanism, sort the encoded patterns and generate the relative weights. To retain the most important feature patterns.

### *1) Top-K Softmax Attention Mechanism*

After the original information is embedded as a high-dimensional dense vector, its expressive power is significantly enhanced. Meanwhile, correspondingly, it also brings problems such as excessive dimension, a large amount of redundant information and high interaction cost. Therefore, it is necessary to compress features at different scales to achieve the goal of reducing the computational load and improving the prediction accuracy. Here, we choose to use the MLP model to generate the encoding weights at the time and channel scales in the order of "time → channel". However, MLP only reduces the dimension through linear transformation and does not distinguish the importance of features, that is, there is no additional attention or gating mechanism to explicitly distinguish the importance of each feature. After this, the Top-k Softmax completes the work of filtering noise and distinguishing the importance of features by sparsifying attention, further supplementing and improving the feature work of the former.

In Li-Net, whether on the time scale or the channel scale, the compression of features is achieved through MLP. The closely connected structure between Top-K Softmax and MLP makes the Logits output by MLP on this scale the sorting basis of Top-K Softmax. This gating mechanism can be formally expressed as the following steps:

#### *a) Initial input and determination of the k value*

The initial input Logits vector is $z = [z_1, z_2, \ldots, z_N] \in \mathbb{R}^N$. After filtering the first k maximum elements ($k \leqslant N$), the final output of Top-k Softmax is $p = [p_1, p_2, \ldots, p_N] \in \mathbb{R}^N$, where p is the dimension of the target feature after MLP compression.

#### *b) Top-k selection*

Select the k elements with the largest values in z to generate the mask vector $m \in \mathbb{R}^N$.

$$m_i = \begin{cases} z_i & if\, z_i \in top_k(z) \\ -\infty & otherwize \end{cases} \quad (4)$$

This operation is implemented through the $top_k(\cdot)$ function, and non-top-k positions are replaced with $-\infty$.

*c) Subset Softmax normalization*

Performing a Softmax operation on the mask vector:

$$p_i = \frac{e^{m_i}}{\sum_{j=1}^{N} e^{m_j}} = \begin{cases} \frac{e^{m_i}}{\sum_{j \in I_k} e^{z_j}} & if\, z_i \in top_k(z) \\ 0 & otherwize \end{cases} \quad (5)$$

Among them, $I_k$ is the set of indices of the Top-k elements.

*2) Time encoding*

In terms of time scale, time series have different quantities of time characteristics from different perspectives. For instance, taking the supply chain scenario as an example, from a date perspective, if measured in hours, purchasing behaviors on working days mostly occur in the morning and evening. When the time scale is measured in days, holidays and weekends will significantly increase people's consumption frequency. Additionally, from the perspective of promotion plans, whether there is a promotion or not will also change over time. Meanwhile, the point of sale itself and the promotion plan also participate in the feature coding work as context vectors between the channel scale and the time scale. The feature encoding work in the time dimension can be formally expressed as the following steps:

*a) The temporal feature tensor to be encoded*

The date embedding tensor of the original time series $T_{hd} = [B, T, D_{date}]$, The point of sale embedded tensor $T_s = [B, T, D_{store}]$. Here, B represents the batch size, T is the historical time step, and $D_d, D_s$ are the vector embedding dimensions of the date and the store, respectively.

*b) Generation of weight matrix*

After concatenating each time feature tensor by time dimension, the time scale feature tensor $T_{pc} = [B, T, (D_d + D_s)]$ is obtained, which is the set of all features on the time scale in the historical time series. After inputting T_pc into the MLP module, the activation function ReLU is used for nonlinear transformation to obtain the initial Logits, and the Logits can obtain the encoding weight matrix of the time scale through Top-k Softmax gating. This step can be expressed as:

$$T_{te} = TopKSoftmax(W_1 \cdot T_{cat-t} + b_1) = [\mathrm{B}, \mathrm{T}, \mathrm{T}']$$

*c) Compression result generation*

After obtaining the weight matrix, we multiply the original time series $\mathrm{T_O} = [\mathrm{B}, \mathrm{C}, \mathrm{T}]$ by the weight matrix tensor T_te, thus completing the compression of the original time series on the time scale and obtaining the tensor T_t, which is formally expressed as follows:

$$\mathrm{T_t} = [\mathrm{B}, \mathrm{C}, \mathrm{T}] \times [\mathrm{B}, \mathrm{T}, \mathrm{T}'] = [\mathrm{B}, \mathrm{C}, \mathrm{T}']$$

*3) Channel encoding*

From the perspective of channel scale, the point-of-sale features and individual item features directly affect the importance of each channel. Since there is an inseparable relationship between the promotion plan and the point-of-sale features, they will also be input as feature tensors here to help the model better select appropriate features at the channel scale. The feature encoding work of the channel dimension can be formally expressed as the following steps:

**Algorithm 3 Non-linear Module**

**Input:** Encoded feature tensor $\mathbf{T_e}$,Choice of block **B**

**Output:** Result of Non-linear Module $\mathbf{R_b}$

**if** B = MLP **then**
 $\mathbf{R_t}$ ←ReLU($\mathbf{T_e} \times \boldsymbol{W_t} + \boldsymbol{b_t}$)while $\boldsymbol{W_t} \in \mathbb{R}^{\mathbf{T'} \times \mathbf{T'}}$
 $\mathbf{R_t}$' ← Transposition($\mathbf{R_t}$)
 $\boldsymbol{R_c}$ ←ReLU($\mathbf{R_t}' \times \mathbf{W}_c + \mathbf{b}_c$)while $\mathbf{W_c} \in \mathbb{R}^{\mathbf{C'} \times \mathbf{C'}}$
 $\boldsymbol{R_b}$ ← Transposition($\boldsymbol{R_c}$)
**elif** B=Transformer **then**
 $\boldsymbol{R_b}$ ← TransformerEncoder($\mathbf{T_e}$)
**return** $\mathbf{R_b}$

*a) The feature tensor of the channel to be encoded*

The original time series tensor $\mathrm{T_t} = [\mathrm{B}, \mathrm{C}, \mathrm{T}']$ encoded only for the channel scale; The embedded tensor of the commodity ID $\mathrm{T_{id}} = [\mathrm{B}, \mathrm{C}, \mathrm{D_{id}}]$ ,Here, $\mathrm{D_{id}}$ represents the commodity embedding dimension.

*b) Generation of weight matrix*

Similarly, this is similar to the time coding work, but the input tensor of the weight matrix here is divided into two parts, namely $\mathrm{T_{ce-1}}$ and $\mathrm{T_{ce-2}}$. First, concatenate the feature tenors of each channel by channel dimension to obtain the channel-scale feature tensor $\mathrm{T_{cat-c}} = [\mathrm{B}, \mathrm{C}, (\mathrm{T}' + \mathrm{T} + \mathrm{P})]$, which is the set of all features at the channel scale in the historical time series. After inputting $T_{cat-c}$ into the MLP module to obtain the initial Logits. This step can be expressed as:

$$\mathrm{T_{ce-1}} = \mathrm{W_2} \cdot \mathrm{T_{cat-c}} + \mathrm{b_2}$$

Furthermore, we obtained $\mathrm{T_{ce-2}}$ through the same operation. Since it is highly similar to the $\mathrm{T_{ce-1}}$ operation, it will not be elaborated here. After adding the two, the coding weight matrix at the channel scale is obtained through Top-k Softmax gating. This step can be formally expressed as:

$$\mathrm{T_{ce}} = \mathrm{TopKSoftmax}(\mathrm{T_{ce-1}} + \mathrm{T_{ce-2}}))$$

*c) Compression result generation*

To facilitate matrix multiplication, here it is necessary to first transpose the weight matrix tensor $\mathrm{T_{ce}}$ and then multiply it by the original time series tensor $\mathrm{T_t}$ encoded on the time scale. Thus, the feature encoding work on both the time scale and the channel scale is completed, obtaining the tensor $\mathrm{T_e}$, which is formally expressed as follows:

$$\mathrm{T_e} = [\mathrm{B}, \mathrm{C}', \mathrm{C}] \times [\mathrm{B}, \mathrm{C}, \mathrm{T}'] = [\mathrm{B}, \mathrm{C}', \mathrm{T}']$$

## *C. Non-linear Module*

To effectively capture the complex nonlinear interaction relationships in the spatiotemporal features after compression and aggregation, we have designed a flexible and efficient nonlinear feature transformation module. This module receives the aggregated feature $\mathrm{T_e}$ generated by spatio-temporal attention and gating mechanism as input. Its goal is to further refine and enhance the feature representation through nonlinear mapping, providing more discriminative information for the final decoding and prediction stage.

*1) Design motivation*

Although the encoder in the previous stage achieved feature aggregation and screening through the MLP module and Top-k Softmax, there may still be high-order nonlinear relationships that have not been fully exploited among its output features. Further, the weights of the encoder output are based on the weights output after feature sorting. However, there are also complex interrelationships among features (such as global dependencies among features), and a simple linear layer or shallow network is difficult to model these complex patterns. Therefore, we introduce an independent deep nonlinear module specifically responsible for performing high-level feature transformations. This module is designed in an "identity mapping" style to ensure consistent input and output dimensions, thus enabling seamless integration into the entire encoder-decoder architecture and significantly enhancing the model's representational ability without disrupting the gradient flow.

*2) Module structure*

We offer two optional sub-module structures to address different computing resource constraints and task complexity requirements, demonstrating the flexibility of model design:

*a) Multi-layer perceptron*

When computing resources are limited or the feature interaction mode is more local, we adopt a lightweight multi-layer perceptron (MLP) block. Its structure is as follows:

$$\mathrm{BLK_{MLP}(X) = Linear_{C''}(ReLU(Linear_{T''}(Transpose(X))))}$$

Among them, $\mathrm{Linear_{C''}}$ and $\mathrm{Linear_{T''}}$ respectively represent the linear transformation layers in the channel dimension and time dimension.

*b) Transformer*

To better model global dependencies and long-range interactions, we offer an alternative solution based on the Transformer encoder. This variant regards the input $\mathrm{T_e}$ as a sequence of length $\mathrm{C'}$, and the feature dimension of each element in the sequence is $\mathrm{T'}$. Its calculation process is:

$$\mathrm{BLK_{Transformer}(X) = TransformerEncoder(X)}$$

*3) Integration of the overall architecture*

This nonlinear module is placed between the encoder and the decoder. Specifically, its input $\mathrm{T_e}$ is the feature aggregated by the sparse attention weights $\mathrm{T_{ce}}$ on the channel dimension. The output $R_b$ transformed by this module is expressed as:

$$\mathrm{R_b = Block}$$

This output is then mapped back to the original feature space through another set of decoder attention weights, and is added to the encoder output $T_{te}$ through residual connection to form the features that are finally passed to the time decoder.

## *D. Feature decoder*

This module adopts an inside-out decoding paradigm, and its design sequence is determined based on the intrinsic semantics of the data and the principle of computational efficiency. First, channel decoding is carried out, aiming to map the high-level abstract features back to the original entity space and determine the specific objects to be predicted. Subsequently, time decoding is carried out, aiming to distribute the entity features to the target prediction time point and generate the final time series prediction result. This sequence simulates the natural decision-making logic of "first determining the entity and then predicting its temporal performance", while mathematically ensuring the highest efficiency of tensor operations. All entities can be multi-step predicted through a single matrix multiplication.

**Algorithm 4 Feature Decoder**

**Input:** Future date embedding tensor $\mathbf{T_{FD}}$,Store embedding tensor $\mathbf{T_s}$, Channel grouping weight $\mathbf{T_{ce-2}}$,, Result of Non-linear Module $\mathbf{R_b}$, The tensor after the temporal features are encoded $\mathbf{T_t}$
**Output:** Result of forecasting $\mathbf{T_{out}}$ with shape of [B,C,P]
1: Logits-2$\in \mathbb{R}^{B\times C\times C'}$ ←MLP( ($\mathbf{T_t}$))
2:$\mathbf{T}_{cd} \in \mathbb{R}^{\mathrm{B\times C\times C'}}$ ←**TopKSoftmax**(Logits-2$\oplus$ $\mathbf{T_{ce-2}}$)
3:preout ←$\mathbf{T}_{cd} \cdot \mathbf{R_b} \oplus \mathbf{T_t}$
4: Logits-3$\in \mathbb{R}^{B\times P\times \mathrm{T'}}$ ←MLP(Concatenate($\mathbf{T_{FD}}, \mathbf{T_s}$))
5: $\mathbf{T_{td}} \in \mathbb{R}^{\mathbf{B\times P\times T'}}$ ←**TopKSoftmax**(Logits-3)
6: $\mathbf{T_{out}}$ ←preout· $\mathbf{T_{td}}$
8:**return** $\mathbf{T_{out}}$

*1) Channel decoding*

Here, an independent yet symmetrical adaptive learning approach is adopted. Although the decoding work is symmetrical to the encoding work, and the input and operations received by the channel decoding weight tensor and the channel encoding weight tensor are the same, their training is independent, and thus the resulting weight matrices are different.

We multiply the channel decoding weight $\mathrm{T_{cd}}$ by the blocknet result $R_b$ to obtain the time series tensor $\mathrm{T_c}$ for decoding only in the channel dimension. Then, by adding the channel decoding output to $\mathrm{T_t}$ element by element with the idea of skip connection, we can obtain the pre-output $\mathrm{T_{pre}}$. This process can be formally expressed as:

$$\mathrm{T_c = T_{cd} \times R_b = [B, C, C'] \times [B, C', T'] = [B, C, T']}$$

$$\mathrm{T_{pre} = T_c + T_t}$$

The purpose of the skip connection is to alleviate vanishing gradients, because if only the last layer is output, it is very likely to cause vanishing gradients, resulting in the loss of some features. The residuals after the jump connection can promote the training of deep networks, facilitate the model's learning of residuals, and at the same time retain the low-level, multi-scale feature information.

*2) time decoding and prediction*

The time decoder is the final stage of the prediction process of this model. Its core responsibility is to map the rich feature representation $\mathrm{T_{pre}}$ obtained after encoding, nonlinear transformation and channel decoding to the future target time series space and generate the final formal prediction result $\mathrm{T_{out}}$. This process is similar to the time coding process. The steps of time scale decoding and the final predicted output can be formally expressed as:

*a) Future time feature tensor*

The generation of time-scale decoding weights requires concatenation of the future date embedding tensor $T_{fd} = [B, P, D_d]$, and the point of sale embedding tensor $T_s = [B, P, D_s]$.

*b) The weighting matrix generation and predict*

After concatenating each future time feature tensor by time dimension, $T_{fc} = [B, P, (D_a + 1 + D_s)]$ is obtained, which is the set of all features on the time scale in the future time series. After inputting $T_{fc}$ into the MLP module in the same way and passing through the activation function ReLU to obtain Logits, the time decoding weight matrix is obtained through Top-K Softmax gating. The final prediction result $\mathrm{T_{out}}$ is obtained by multiplying the pre-output matrix $\mathrm{T_{pre}}$ decoded at the channel scale with the transposed time-decoded weight matrix $T_{td}$. This process can be formally expressed as:

$$T_{td} = TopKSoftmax(W_1 \cdot T_{fc} + b_1) = [\mathrm{B}, \mathrm{P}, \mathrm{T}'']$$

$$\mathrm{T_{out}} = [\mathrm{B}, \mathrm{C}, \mathrm{T}'] \times [\mathrm{B}, \mathrm{T}'', P] = [\mathrm{B}, \mathrm{C}, \mathrm{P}]$$

## V. Experiment

### A. Experimental Setup

*1) Datasets*

To comprehensively evaluate the performance of the Li-Net framework we proposed and ensure a fair comparison with existing baseline models, we conducted experiments on five publicly available and widely recognized time series prediction benchmark datasets from different domains. These datasets vary in time series dynamics, sampling frequencies, and data characteristics, providing a comprehensive test platform for evaluating the generalization ability of prediction models.

**ETT** dataset is a key benchmark for long-term time series prediction, recording the load and oil temperature data of power transformers in two different power stations in China. We used its subsets of every hour and every 15 minutes. Its high-frequency sampling and strong long-range dependence pose significant challenges for the model to capture temporal evolution patterns. This dataset contains the hourly electricity consumption (unit: kWh) of 321 customers from 2012 to 2014. The prediction task is to predict the power load of all future customers, making it a classic multivariate prediction problem. The data shows strong daily and weekly periodicity, as well as correlations among different customers.

**Weather** dataset contains 21 meteorological indicators recorded at a certain meteorological station in Germany within a year, with records made every 10 minutes. The challenge of this data lies in modeling the complex nonlinear interactions among these variables and making precise predictions under highly non-stationary conditions.

**Traffic** dataset describes the hourly occupancy rates of 862 lanes on San Francisco's highways from 2015 to 2016. As a typical spatio-temporal prediction problem, this data features complex short-term patterns and long-term trends, requiring the model to effectively capture temporal correlations and cross-correlations among different paths.

**M5** dataset is provided by Walmart and is a large-scale hierarchical sales forecasting dataset. It contains daily sales data of 3,049 products from 10 stores and is organized in a complex spatio-temporal hierarchy. Predicting this data requires modeling multi-scale seasonality (daily, weekly, annual) and handling the harmonization problem of prediction results at different aggregation levels (products, stores, categories, etc.), which poses a huge challenge to the scalability and accuracy of the model.

*2) Metrics*

To comprehensively and fairly evaluate the overall performance of the Li-Net framework and its baseline model, we selected evaluation metrics from three core dimensions: prediction accuracy, computational efficiency, and deployment feasibility. This multi-angle evaluation strategy aims to ensure that the proposed model is not only competitive in prediction accuracy, but also has advantages in practicality and resource efficiency.

From the perspective of prediction accuracy, we adopt MAE and MSE as the core indicators for evaluating prediction accuracy. MAE can visually measure the average deviation of predicted values, while MSE is more sensitive to larger errors. The combination of the two can comprehensively evaluate the robustness and accuracy of the model under different error performances [41].

From the perspective of computational efficiency, we recorded the running time of the model to measure the computational efficiency of its training and inference processes. This indicator directly reflects the response speed and processing capacity of the model in practical applications, and is crucial for scenarios that require high-frequency prediction or real-time decision-making [42].

From the perspective of deployment feasibility, we have introduced the size of occupied memory and model size as key evaluation indicators. The size of memory occupied reflects the demand for hardware resources during the model's operation and is a key factor determining whether the model can be deployed on resource-constrained devices [43]. The size of the model represents its complexity and storage cost. Smaller models are easier to transfer, store and deploy [44]. These two indicators jointly evaluated the practicability and engineering potential of the model [45].

Through the comprehensive consideration of the above multi-dimensional indicators, our experimental evaluation not only focuses on the theoretical performance of the model, but also pays attention to its overall performance in practical application scenarios, because an excellent model should achieve the best balance between accuracy and efficiency [46].

*3) Baselines*

To comprehensively evaluate the performance of the Li-Net framework we proposed, we compared it with a series of representative advanced models. The selected baselines cover time series prediction models based on different architectures, including classic recurrent neural networks, convolutional networks, modern Transformer variants based on attention mechanisms, and specialized time series hybrid models, ensuring the breadth and fairness of the comparison.

iTransformer[15] : This model represents a significant reflection on the Transformer architecture in the field of time series prediction in recent times. The core idea is to treat all the variables at each time point as a token and calculate attention on the variable dimension rather than the traditional time dimension.

TimeMixer[47] : TimeMixer is a lightweight model based on MLP. It achieves information interaction by separately modeling the endogenous and exogenous patterns of time series and introducing efficient hybrid modules.

PatchTST[48] : The PatchTST model divides the time series into patches and applies Transformer to model the Patch sequence. This method effectively increases the semantic information of the input sequence and reduces the sequence length. It is one of the most advanced Transformer-based prediction models currently available and serves as the core baseline for our comparison.

**Table I: Overall performance comparisons**

| Datasets | | Li-Net | | TimeMixer | | TFT | | PatchTST | | iTransformer | | TCN | |
|---|---|---|---|---|---|---|---|---|---|---|---|---|---|
| | | MAE | MSE | MAE | MSE | MAE | MSE | MAE | MSE | MAE | MSE | MAE | MSE |
| ETTh2 | **96** | **0.282** | **0.1641** | 0.3406 | 0.2888 | 0.3784 | 0.3620 | 0.3523s6 | 0.296 | 0.349 | 0.2989 | 1.4768 | 3.1767 |
| | **192** | **0.3092** | **0.1945** | 0.3947 | 0.3771 | 0.4648 | 0.5011 | 0.3993 | 0.377 | 0.396 | 0.3786 | 5.2776 | 2.0467 |
| | **336** | **0.3233** | **0.2028** | 0.4394 | 0.4271 | 0.4746 | 0.5128 | 0.4419 | 0.429 | 0.433 | 0.4303 | 2.0550 | 5.7964 |
| | **720** | **0.3553** | **0.2408** | 0.4685 | 0.4686 | 0.4950 | 0.5364 | 0.4627 | 0.443 | 0.442 | 0.4248 | 2.0932 | 6.9208 |
| ETTm2 | **96** | **0.2284** | **0.1131** | 0.2610 | 0.1788 | 0.2943 | 0.2131 | 0.2717 | 0.183 | 0.268 | 0.1843 | 0.4922 | 0.4323 |
| | **192** | **0.2612** | **0.1436** | 0.3065 | 0.2474 | 0.3364 | 0.2770 | 0.3061 | 0.244 | 0.313 | 0.2541 | 0.6756 | 0.7621 |
| | **336** | **0.2891** | **0.1723** | 0.3424 | 0.3016 | 0.3689 | 0.3352 | 0.3487 | 0.311 | 0.350 | 0.3144 | 0.8880 | 1.2207 |
| | **720** | **0.3231** | **0.2149** | 0.3988 | 0.4021 | 0.4268 | 0.4428 | 0.4173 | 0.432 | 0.406 | 0.4115 | 1.7267 | 4.3460 |
| Electricity | **96** | **0.275** | **0.1757** | 0.2762 | 0.1764 | 0.3052 | 0.2039 | 0.2866 | 0.189 | 0.370 | 0.2885 | 0.4309 | 0.3607 |
| | **192** | **0.2867** | **0.1882** | 0.2879 | 0.1896 | 0.3123 | 0.2122 | 0.2935 | 0.196 | 0.394 | 0.3136 | 0.4032 | 0.3241 |
| | **336** | **0.3029** | **0.2022** | 0.3064 | 0.2081 | 0.3284 | 0.2295 | 0.3107 | 0.214 | 0.413 | 0.3361 | 0.4429 | 0.3844 |
| | **720** | **0.3369** | **0.2434** | 0.3413 | 0.2508 | 0.3497 | 0.2578 | 0.3447 | 0.257 | 0.344 | 0.2672 | 0.4094 | 0.3414 |
| Weather | **96** | 0.2407 | **0.1757** | 0.2526 | 0.1777 | **0.2357** | 0.1917 | 0.2444 | 0.179 | 0.240 | 0.1988 | 0.293 | 0.2123 |
| | **192** | **0.2865** | **0.22** | 0.3037 | 0.2278 | 0.2952 | 0.2661 | 0.2927 | 0.229 | 0.275 | 0.2442 | 0.3346 | 0.2622 |
| | **336** | **0.3142** | 0.2676 | 0.3208 | **0.2630** | 0.3304 | 0.3236 | 0.3387 | 0.288 | 0.326 | 0.2958 | 0.3945 | 0.3435 |
| | **720** | **0.3676** | **0.3389** | 0.3772 | 0.3432 | 0.3851 | 0.4119 | 0.3976 | 0.370 | 0.375 | 0.3677 | 0.4211 | 0.4102 |
| Traffic | **96** | **0.3171** | **0.3482** | 0.3337 | 0.4712 | 0.3470 | 0.5921 | 0.4176 | 0.672 | 0.550 | 0.8986 | 0.4178 | 0.7362 |
| | **192** | **0.3317** | **0.3676** | 0.3346 | 0.4851 | 0.3461 | 0.5989 | 0.3986 | 0.627 | 0.562 | 0.9104 | 0.4483 | 0.7506 |
| | **336** | **0.3333** | **0.3745** | 0.3416 | 0.4999 | 0.3476 | 0.6049 | 0.4013 | 0.636 | 0.572 | 0.9361 | 0.4643 | 0.7801 |
| | **720** | **0.3462** | **0.3942** | 0.3570 | 0.5378 | 0.3530 | 0.6294 | 0.4181 | 0.675 | 0.587 | 0.9802 | 0.4883 | 0.8218 |
| M5 | **96** | **0.2051** | **0.2014** | 0.2085 | 0.3224 | 0.2244 | 0.2639 | 0.2323 | 0.357 | 0.298 | 0.4258 | 0.3309 | 1.095 |
| | **192** | **0.2309** | **0.2464** | 0.2321 | 0.4213 | 0.2360 | 0.2589 | 0.3071 | 0.494 | 0.357 | 0.5451 | 0.4119 | 2.6692 |
| | **336** | 0.2976 | 0.326 | 0.2996 | 0.5290 | **0.1051** | **0.2164** | 0.3331 | 0.562 | 0.404 | 0.6346 | 0.5622 | 2.0633 |
| | **720** | 0.3399 | 0.3695 | 0.3540 | 0.5902 | **0.1441** | **0.2364** | 0.3827 | 0.642 | 0.429 | 0.6925 | 0.6498 | 1.7902 |

TCN[49] : TCN utilizes dilated causal convolution to process sequences in parallel, combining the efficiency of convolutional networks with the ability to capture long-range dependencies. It is a successful application of convolutional neural networks in time series modeling, and we choose it as the representative baseline of the convolutional architecture.

TFT[50] : TFT is a powerful model specifically designed for probabilistic prediction. It can simultaneously model known time-varying inputs, static variables, and learn long-term and short-term dependencies in sequences.

*4) Implementation Details*

This experiment was trained using the AdamW optimizer. This optimizer was chosen because of its effectiveness in handling weight decay regularization and its outstanding performance in training deep neural networks. We use the PyTorch 2.6.0 framework that supports CUDA 12.4 and Python 3.11.11 to implement the Li-Net framework we proposed. All experiments were conducted on a server equipped with an NVIDIA L40S GPU. The dataset is divided into a training set, a validation set and a test set in a ratio of 60% : 20% : 20%. We adopted 16 as the batch size for model training and evaluation, set the epoch to 10, and the patience to 3. To conduct a comprehensive and fair comparison, we selected all baseline models from the widely recognized Tsinghua time Series library, which ensured that our experimental comparisons were carried out under consistent and reproducible conditions.

*B. Overall performance*

Li-Net demonstrated outstanding overall performance in multivariate time series prediction tasks, significantly outperforming other benchmark models. In terms of both MAE and MSE metrics, Li-Net ranked first or second in 23 out of 32 Settings across 8 datasets. This not only validates its status as an SOTA prediction model but also highlights its practical value in industrial deployment - the combination of low error and high efficiency makes it suitable for real-time prediction scenarios. In terms of average indicators, Li-Net's MAE and MSE are both at the lowest. The average MAE was 0.3443 and the average MSE was 0.2993, which were much lower than those of iTransformer (average MAE 0.4064, average MSE 0.4583) and PatchTST (average MAE 0.3673, average MSE 0.3814). This advantage indicates that Li-Net has a significant edge in reducing prediction bias and error variance, enabling it to capture the dynamic patterns of time series more accurately while maintaining efficient computational characteristics.

**Table II: Ablation Analysis**

| Datasets | Linet | | Linet-Softmax | | Linet-MLP | | Linet-Primitive | |
|---|---|---|---|---|---|---|---|---|
| | MAE | MSE | MAE | MSE | MAE | MSE | MAE | MSE |
| ETTm2_96 | **0.2284** | **0.1131** | 0.236 | 0.1175 | 0.5415 | 0.4847 | 0.2418 | 0.1215 |
| ETTm2_336 | **0.2891** | **0.1723** | 0.2983 | 0.1814 | 0.5542 | 0.5171 | 0.3081 | 0.1937 |
| Electricity_96 | **0.275** | **0.1757** | 0.2804 | 0.1882 | 0.8225 | 0.9718 | 0.2816 | 0.1806 |
| Electricity_336 | **0.3029** | **0.2022** | 0.3071 | 0.2076 | 0.8205 | 0.9588 | 0.3137 | 0.215 |

## *C. Memory usage and efficiency*

We conducted a comprehensive assessment of the Li-Net framework from the perspective of computational efficiency, including training and inference time, memory usage, and model size. The results show that Li-Net significantly outperforms other baseline models in terms of computational efficiency while maintaining high prediction accuracy, demonstrating its practicality and deployability in resource-constrained environments.

### *1) Time efficiency*

As shown in Fig.2a and Fig.2b, Li-Net performs well in terms of training and testing duration, especially on long sequences and large-scale datasets. For instance, on the ETTh2 dataset, the training duration of Li-Net at different time steps (96-720) was 38.25 seconds, 60.28 seconds, 47.98 seconds, and 50.24 seconds respectively, which was significantly lower than that of many baseline models, such as TimeMixer and PatchTST. In terms of test duration, Li-Net also leads, with a test time of only 0.4 to 0.56 seconds on ETTh2, while other models such as TFT and PatchTST often have test times exceeding 1 second, and even reach tens of seconds on the Traffic dataset. This high efficiency stems from Li-Net's sparse attention mechanism and multi-scale projection design. By dynamically compressing time and channel dimensions, it reduces computational complexity and avoids the quadratic overhead of traditional Transformers.

### *2) Memory usage*

As shown in Fig.2c and Fig.2d ,The memory usage of Li-Net during training and testing is also significantly lower. On the Traffic dataset, the training memory usage of Li-Net was only 41.17MB to 167.18MB, and the test memory usage was also 41.56MB to 166.19MB. In contrast, the memory usage of other models such as TFT and TimeMixer often exceeded 100MB. Even on the Electricity dataset, it reaches several thousand MB. On the Electricity dataset, the training memory usage of Li-Net was only 27.71 MB to 88.47 MB, while the memory usage of iTransformer and PatchTST both exceeded 400 MB. This low memory requirement benefits from Li-Net's encoder-decoding structure, which filters key features through Top-K attention and reduces the storage of intermediate states, thereby lowering memory pressure.

### *3) Model size*

As shown in Fig.2e, Li-Net demonstrates outstanding parameter efficiency. Its compact design ensures extremely low storage requirements across all datasets. As shown in Table X, the average model size of Li-Net is only 0.5MB, which is much lower than that of other baseline models, especially TFT (26.8MB) and TCN (0.93MB). This advantage is more pronounced on specific datasets: on the ETTm2 dataset, the model size of Li-Net is only 0.5MB to 5.65MB, while the model sizes of PatchTST and TFT typically exceed 20MB. On the Traffic and M5 datasets, the model size of TFT is even close to 1GB, while the maximum model size of Li-Net is only 7.31MB.

This remarkable parameter efficiency stems from the core design concept of the Li-Net architecture: achieving parameter sharing and compression through multimodal embedding and sparse attention mechanisms. Specifically, the Top-K Softmax attention mechanism dynamically screens key features, avoiding parameter explosion in the fully connected layer in traditional attention mechanisms. Meanwhile, the multi-scale projection framework realizes the intelligent compression of features during the encoder-decoding process, retaining only the most discriminative information. The unified embedding layer design reduces model redundancy through parameter reuse.

## *D. Ablation Analysis*

To comprehensively verify the effectiveness of each core component in the Li-Net framework, we designed a systematic ablation experiment and conducted a comprehensive evaluation on the ETTh2 and Electricity datasets. The reason for choosing these two datasets is that they can cover the key challenges in multi-channel time series prediction: The ETTh2 dataset has the characteristic of long sequence dependence and can effectively test the model's ability to capture the dynamics of the time dimension; The Electricity dataset, on the other hand, features high dimensionality and can test the feature screening and fusion performance of models under complex variable associations. We selected 96 and 336 time steps on the ETTh2 dataset, representing short-term and long-term prediction scenarios respectively, and 96 and 336 time steps on the Electricity dataset to verify the model's performance under different dimensional complexities. This choice ensures that the ablation experiment can cover key prediction scenarios while maintaining manageable computational costs.

We designed three ablation experimental variants to isolate and analyze the contribution of each key component:

- **Li-Net-softmax**: A full attention mechanism that replaces the original Top-K Softmax attention with standard Softmax attention to verify the advantages

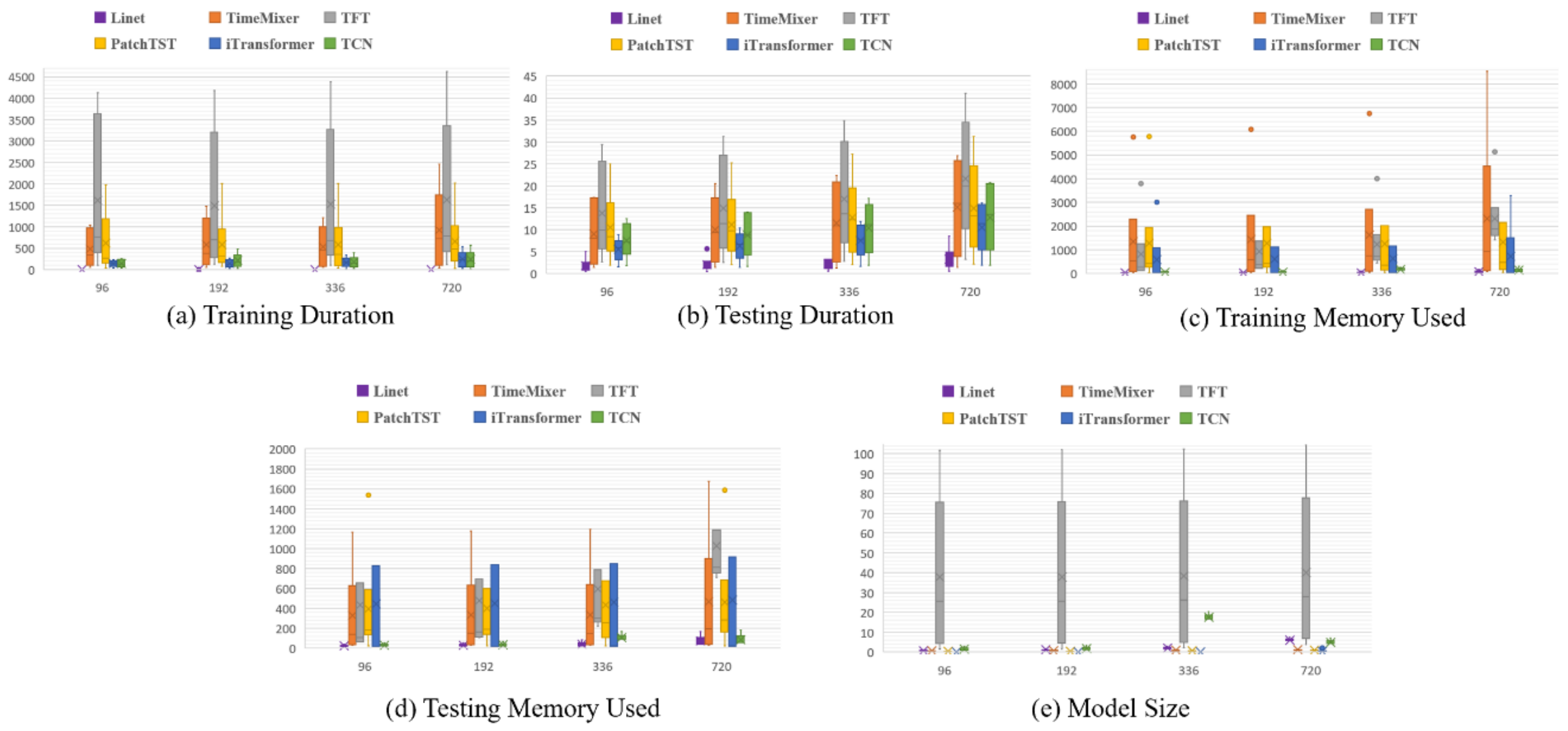

(a) Training Duration (b) Testing Duration (c) Training Memory Used

(d) Testing Memory Used (e) Model Size

**Figure 2: Memory usage and efficiency**

of sparse attention in reducing computational complexity and maintaining accuracy.

- **Li-Net-Primitive**: Remove all multimodal embeddings and use only the original time series data input to quantify the improvement effect of multimodal information on predictive performance.
- **Li-Net-MLP**: Replace the complete encoder-decoder framework with a three-layer MLP and map the fused features directly to the predicted output to evaluate the necessity of the multi-scale projection architecture in feature compression and reconstruction.

The experimental results show that the existing architecture of Li-Net is highly optimized and necessary, and each core component makes an irreplaceable contribution to the final performance. The experimental results of the "Li-Net-Primitive" variant most clearly reveal the key role of multimodal information. In all experimental setups, the performance of this variant that only uses the original time series data lags significantly behind that of the full version of Li-Net. In all four tasks of ETTm2 and Electricity, the MAE and MSE of Li-Net were comprehensively lower than those of the variants using full attention. The Top-K Softmax attention mechanism effectively filters out noise interference by dynamically screening key features. While maintaining high prediction accuracy, it avoids the huge quadratic computational overhead of the standard attention mechanism. This proves the rationality of the design in pursuing the optimal balance between computational efficiency and model performance. The performance of the "Li-Net-project" variant has experienced a catastrophic decline, and its various indicators are far inferior to those of all other model variants, fully demonstrating the importance of complex structures. The multi-scale projection, feature compression and reconstruction mechanism implemented by the encoder-decoder structure is the fundamental reason why Li-Net has powerful representation capabilities. This design is indispensable for extracting effective information from high-dimensional inputs and modeling complex nonlinear patterns.

## *E. Parameter Studies*

To deeply understand the impact of each key hyperparameter in the Li-Net framework on performance and provide configuration guidance for actual deployment, we conducted a parameter sensitivity analysis of the system on the ETTm2 and Electricity datasets, focusing on four core dimensions: sparse attention retention rate, feature compression level, nonlinear module selection, and embedding dimension.

### *1) Sparse Attention Retention Analysis*

As shown in Fig3a and Fig3g, we first evaluated the impact of the attention retention rate in the time dimension on the model performance. Under the conditions of maintaining the channel retention rate at 0.7, the time compression level at 2, the nonlinear module at MLP, and the embedding dimension at 32, we tested the performance variation of the time retention rate within the range of {0.1, 0.3, 0.5, 0.7, 0.9}.

Analysis shows that a moderate time retention rate (0.3-0.5) performs best in most scenarios. Specifically, on the ETTm2 dataset, when the time retention rates were 0.3 and 0.5, the model achieved MAE of 0.2298 and 0.2298 respectively in the 96-step prediction task, significantly outperforming the extreme Settings. In the 336-step prediction, the MAE corresponding to a time retention rate of 0.3 is 0.2846, which is superior to other configurations.

As shown in Fig3b and Fig3h, we then evaluated the impact of the channel dimension attention retention rate. Under the conditions of a fixed time retention rate of 0.5, a time compression level of 2, a channel compression level of 2, a nonlinear module of MLP, and an embedding dimension of 32, the performance of the channel retention rate within the range of {0.1, 0.3, 0.5, 0.7, 0.9} was tested.

The experimental results show that the influence of channel retention rate on model performance exhibits dataset dependence. On the ETTm2 dataset, the performance differences under different channel retention rates are relatively small, and the optimal MAE fluctuates between 0.2295 and 0.2396. However, on the Electricity dataset, a lower channel retention rate (0.1-0.3) performed better, especially in the 336-step prediction. The MAE corresponding to a channel retention rate of 0.1 was 0.304, which was significantly better than 0.3122 at 0.9. This indicates that in high-dimensional channel datasets, moderate channel sparsity helps filter out redundant variable interactions and enhance the generalization ability of the model. It is recommended to adopt a retention rate of 0.1 to 0.3 in the channel dimension.

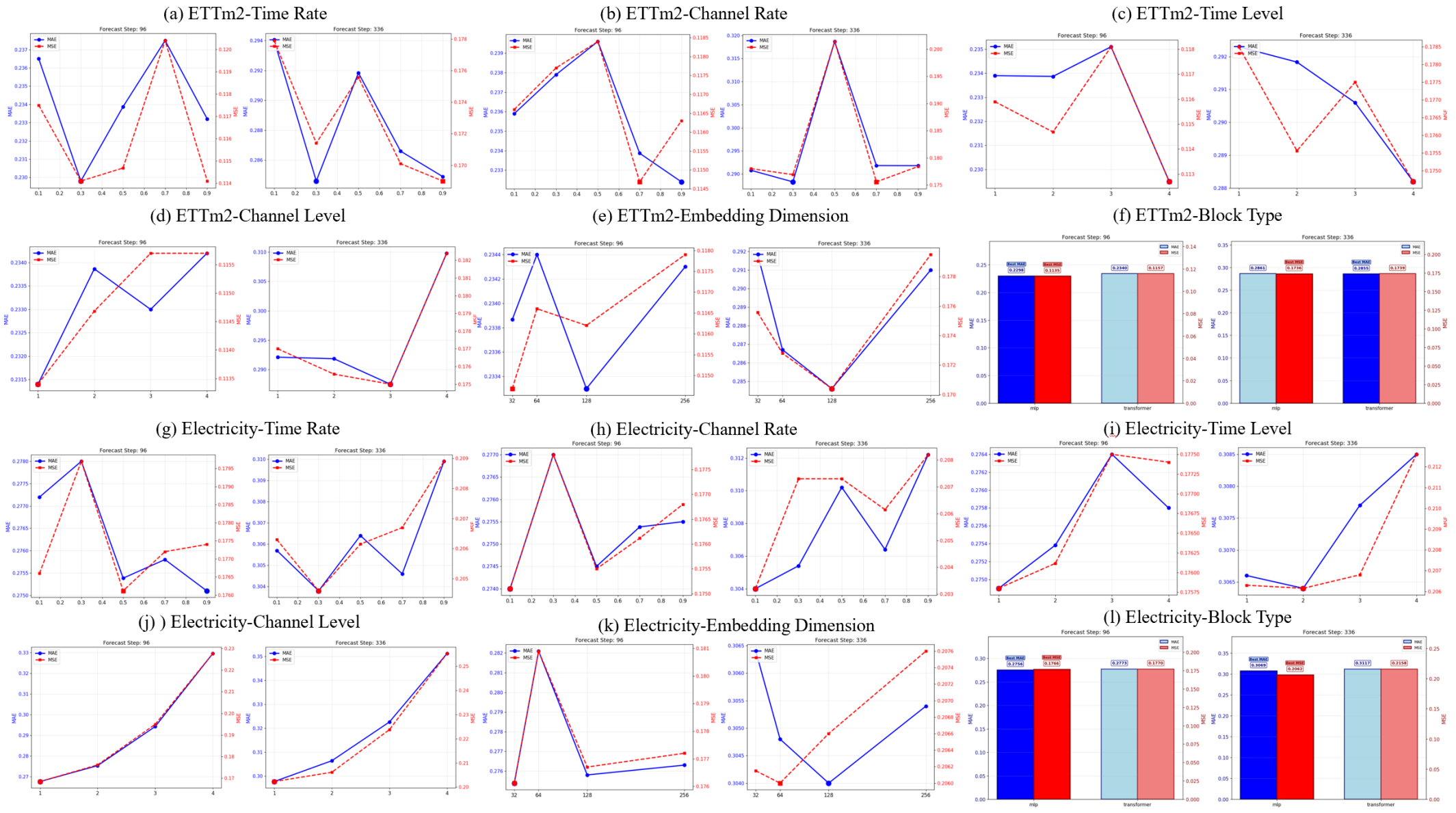

**Figure 3: Parameter Analysis**

*2) Impact of feature compression level*

As shown in Fig3c, Fig3i, Fig3d, and Fig3j, we further analyzed the influence of time and channel compression level. The experiment maintained the time retention rate of 0.5, the channel retention rate of 0.7, the MLP of the nonlinear module, and the embedding dimension of 32 unchanged.

Time compression level analysis shows that on the ETTm2 dataset, a higher level of time compression (level=4) achieved the best performance in 96-step predictions (MAE=0.2295), indicating that moderate time series aggregation is helpful for capturing long-term patterns. On the Electricity dataset, the performance differences among different time compression levels are relatively small, and the optimal configuration occurs at level=1.

Channel compression level analysis reveals more significant impacts. On the Electricity dataset, the model performs best when the channel compression level is 1, but as the compression level increases to 4, the performance drops significantly. This indicates that in high-dimensional channel datasets, excessive channel compression will result in the loss of key variable interaction information. Comprehensive suggestion: Set the time compression level to 4 and the channel compression level to 1.

*3) Multimodal embedding dimension analysis*

As shown in Fig3e and Fig3k, the embedding dimension analysis tested four dimension levels: {32, 64, 128, 256}. The results show that a smaller embedding dimension performs best in most scenarios, achieving 0.2339 MAE on ETTm2 and 0.2754 MAE on Electricity. As the dimension increased, the performance did not show a significant improvement; instead, it declined in some cases, while the number of model parameters increased significantly. Embedding dimension 32 is determined as the optimal balance point between computational efficiency and expressive power.

## VI. Conclusion

This paper proposes Li-Net, a novel and efficient architecture for multi-channel time series prediction. By introducing the Top-K sparse attention mechanism and the multi-scale projection framework, Li-Net significantly reduces the computational and memory overhead caused by the traditional attention mechanism while maintaining high prediction accuracy.

Extensive experiments on multiple public time series datasets have shown that Li-Net outperforms current mainstream baseline models in both prediction accuracy and computational efficiency, demonstrating superior comprehensive performance. The ablation experiments and parameter sensitivity analysis of the system further verified the necessity and design rationality of each core component.

In the future, we will explore extending Li-Net to more challenging scenarios, such as online learning, multi-granularity time series prediction, and cross-domain time series transfer learning, to further enhance its applicability and generalization ability in actual industrial environments.